\documentclass{article}
\usepackage{spconf,amsmath,graphicx,amssymb}

\ninept
\usepackage{algorithm,amsfonts}
\usepackage{float}
\usepackage{algpseudocode}
\usepackage[usenames,dvipsnames]{color} 
\usepackage{tabularx} 
\usepackage{cite}
\usepackage{caption}
\usepackage{subcaption}
\input{latex_resources/mysymbol.sty}
\input{latex_resources/juan_defs.sty}
\usepackage{multirow}
\title{Training Graph Neural Networks on Growing Stochastic Graphs}
\name{Juan Cervi\~no$^1$, Luana Ruiz$^2$, and Alejandro Ribeiro$^1$ \thanks{Support by NSF CCF 1717120, ARL DCIST CRA under Grant W911NF-17-2-0181 and Theorinet Simons.}}
\address{$^1$Department of Electrical and Systems Engineering, University of Pennsylvania, Philadelphia, USA\\
$^2$Simons-Berkeley Institute}
\begin{document}
%\ninept
%
\maketitle
\begin{abstract}
Graph Neural Networks (GNNs) rely on graph convolutions to exploit meaningful patterns in networked data. Based on matrix multiplications, convolutions incur in high computational costs leading to scalability limitations in practice. To overcome these limitations, proposed methods rely on training GNNs in smaller number of nodes, and then transferring the GNN to larger graphs. Even though these methods are able to bound the difference between the output of the  GNN with different number of nodes, they do not provide guarantees against the optimal GNN on the very large graph. In this paper, we propose to learn GNNs on very large graphs by leveraging the limit object of a sequence of growing graphs, the graphon. We propose to grow the size of the graph as we train, and we show that our proposed methodology -- \textit{learning by transference} -- converges to a neighborhood of a first order stationary point on the graphon data. A numerical experiment validates our proposed approach.
\end{abstract}
\section{Introduction}
\label{sec:intro}

Graph Neural Networks (GNNs) are deep convolutional architectures for non-Euclidean data supported on graphs \cite{kipf2016semi,gama2018convolutional}. GNNs are formed by a succession of layers, each of which is composed of a graph convolutional filter and a point-wise non-linearity. In practice, GNNs have shown state-of-the-art performance in several learning tasks \cite{wang2019dynamic,bronstein2017geometric}, and have found applications in biology \cite{NIPS2017_f5077839, NIPS2015_f9be311e,pmlr-v70-gilmer17a}, recommendation systems \cite{fan2019graph,tan2020learning,ying2018graph} and robotics \cite{qi2018learning,gama2020decentralized}. In theory, part of their success is credited to their stability to graph perturbations \cite{gama2018convolutional}, the fact that they are invariant to relabelings \cite{chen2019equivalence,8911416}, their expressive power \cite{xu2018powerful,kanatsoulis2022graph}, and their generalization capabilities \cite{maskey2022stability}.

The ability of GNNs to exploit patterns in network data is largely due to their graph convolutional layers. Graph convolutional filters leverage the graph diffusion sequence to extract features that are shared across the graph \cite{gama2018convolutional}. However, these filters rely on computing matrix-vector multiplications, which become computationally costly when the number of nodes is large. On the other hand, the number of parameters in a GNN is independent of the number of nodes due to its local parametrization, which motivates training GNNs on small graphs and then deploying them on larger graphs.
%,i.e. to transfer the weights from a small graph to a large one. 

Several works have been proposed which exploit this transferability property of the GNN \cite{ruiz2020graphonTransferability,ruiz2021transferability,ruiz2022machine}. Generally speaking, these works upper bound the output difference between two GNNs with the same parameters supported on graphs with different number of nodes. They show that if the graphs belong to a family of graphs modeled by a graphon \cite{lovasz2012large}, as the number of nodes increases the so-called \textit{transferability error} decreases. But while this result is useful, it does not guarantee than the optimal parameters achieved by training the GNN on the small graph are optimal, or sufficiently close to optimal, on the large graph.
%does not allow us to claim that the optimal GNN trained with a small number of nodes will perform as well as the optimal GNN trained on a very large graph. 
This subtle albeit crucial observation is what motivates 
our work.
%our line of work.

%In this paper, we propose to tackle both problems previously mentioned; (i) the computational cost of deploying GNNs on very large graphs, as well as (ii) obtaining optimal GNNs on the very large graphs. 
%In this work we 
More specifically, we leverage the transferability properties of GNNs to introduce a method that trains GNNs by successively increasing the size of the graph during the training process. We call this procedure \textit{learning by transference}. We prove that the learning directions (gradients) on the graphon and on the graph are aligned as long as the number of nodes in the GNN is sufficiently large. Under a minimum graph size at each epoch, we show that our method converges to a neighborhood of the first order stationary point on graphon by taking gradient steps on the graph. We benchmark our procedure in a multi-agent system problem, where a GNN is trained to learn a decentralized control policy for agents communicating via a proximity graph. 

\noindent \textbf{Related work.} This work extends upon previous works that consider the nodes to be sampled from a regular partition of the graphon \cite{cervino2021increase}. In this work, we consider the nodes to be uniformly sampled, which not only provides a more general modeling of GNNs, but also more closely correlates with graph signals seen in practice. This more general sampling strategy induces further sampling errors (in addition to the edge sampling error) \cite{ruiz2021transferability}, which worsen the approximation of graphons and graphon data by graphs and graph data thus requiring additional theoretical considerations. 

\section{Graph and Graphon Neural Networks}
A graph is represented by the triplet $\bbG_n=(\ccalV,\ccalE,W)$, where $\ccalV,|\ccalV|=n$ is the set of nodes, $\ccalE\subseteq\ccalV\times\ccalV$ is the set of edges, and $W:\ccalE\to \reals$ is a map assigning weights to each edge. Alternatively, we can represent the graph $\bbG_n$ by its graph shift operator (GSO) $\bbS_n\in\reals^{n\times n}$. Examples of GSO are the adjecency matrix $\bbA$, the graph laplacian $\bbL=\mbox{diag}(\bbA\bbone)-\bbA$, to name a few. 

Data on graphs is defined as a graph signal $\bbx=[x_1,\dots,x_n]$, where the $i$th component of $\bbx$ corresponds to the value of the signal on node $i$. A graph signal $\bbx$ can be aggregated through the graph, by applying the GSO $\bbS_n$ as follows, 
\begin{align}
    \bbz = \bbS_n \bbx_n.
\end{align}
Intuitively, the value of signal $\bbz$ at coordinate $i$ is the weighted average of the information present in its one hop neighborhood, i.e. $z_i=\sum_{j\in\ccalN(i)} [\bbS_n]_{ij}x_j$. We can construct $k$-hop diffusion by applying $k$ power of $\bbS_n$.
We define the \textit{graph convolutions}, as a weighted average over the powers of $\bbS_n$. Explicitly, by letting the coefficients be $\bbh=[h_0,\dots,h_{K-1}]$, we define the graph convolution as,
\begin{align}
\bby_n=\bbh_{* \bbS_n}\bbx_n = \sum_{k=0}^{K-1} h_k \bbS_n^k\bbx_n
\end{align}
where $\bbx_n,\bby_n$ are graph signals, and $\bbh_{* \bbS_n}$ denotes the convolution operator with GSO $\bbS_n$. 

In the case of undirected graphs, $\bbS_n/n$ becomes symmetric, admitting a spectral decomposition $\bbS_n=\bbV \bbLambda \bbV^T$, where the columns of $\bbV$ are the eigenvectors of $\bbS_n$, and $\bbLambda$ is a diagonal matrix with $-1\leq \dots \leq \dots 1$. Since the eigenvector of $\bbS_n$ for a basis of $\reals^n$, we can project filter $\bby_n$ into this basis to obtain,
\begin{align}\label{eqn:graph_spectral}
    h(\lambda) = \sum_{k=0}^{K-1} h_k \lambda^k
\end{align}
Note that $h(\lambda)$ only depends on $h_k$, and on the eigenvalues of the GSO. By the Cayley-Hamilton theorem, convolutional filters may be used to represent any graph filter with spectral representation $h(\lambda)=f(\lambda)$, where $f$ is analytic.

\textit{Graph Neural Networks} are layered architectures, composed of graph convolutions followed by point-wise non-linearities. Formally, introducing a point-wise non-linearity $\rho$, and by stacking all the graph signals at layer $l$ in a matrix $\bbX_l=[\bbx_{nl}^1,\dots,\bbx_{nl}^{F_l}]\in\reals^{n\times F_l}$, where $F_l$ indicates the $F_l$ features at layer $l$. Notice that for $l=0$, the input matrix is a concatenation of the input signal, i.e. $\bbX_0=[\bbx_{n},\dots,\bbx_{n}]$. We can define the $l$ layer of a GNN as, 
\begin{align}
    \bbX_{l} = \rho \bigg(\sum_{k=1}^K \bbX_n^k \bbX_{l_1}\bbH_{lk} \bigg),
\end{align}
where matrix $\bbH_{lk}\in\reals^{F_{l-1}\times F_l}$ represents the $k$ coefficient of the graph convolution. By grouping all the learnable parameters we can obtain a more succint representation of the GNN as $\phi(\bbx;\ccalH,\bbS)$, with $\ccalH=\{\bbH_{lk}\}_{lk}, 1\leq l \leq L$, and $0\leq k \leq K-1$.

\subsection{Graphon Neural Networks}
Graphons are the limit object of a converging sequence of dense undirected graphs. Formally, a graphon is a symmetric, bounded, and measurable function $\bbW:[0,1]^2\to[0,1]$. Sequences of dense graphs converge to a graphon in the sense that the densities of adjacency preserving graph motifs converge to the densities of these same motifs on the graphon \cite{ruiz2020graphonTransferability,lovasz2012large}.

Analogously to graph signals, we can define a graphon signal as  a function $X\in L^2([0,1])$. A graphon signal can be diffused over the graphon by applying the linear integrator operator given by, 
\begin{align}
    T_\bbW X(v)=\int_0^1 \bbW(u,v)X(v)du
\end{align}
which we denote graphon shift operator (WSO). Since graphon $\bbW$ is bounded and symmetric, $T_\bbW$ is a Hilbert-Schmidt and self adjoint operator \cite{lax02-functional}. Therefore, we can express $\bbW$ by its eigen decomposition, $\bbW(u,v)=\sum_{i\in\integers \smallsetminus 0}\lambda_i\phi_i(u)\phi_i(v)$, where $\lambda_i$ is the eigenvalue associated with eigenfunction $\phi_i$. The absolute value of the eigenvalues is bounded by $1$, and the eigenfunctions form an orthogonal basis of $L^2([0,1])$. 
Utilizing the graphon shift, we define the graphon convolution with parameters $\bbh=[h_0,\dots,h_{K-1}]$, as
\begin{align}\label{eqn:graphon_filter}
    Y=T_\bbh X = \bbh_{*\bbW} X =\sum_{k=0}^{K-1}h_k(T_\bbW^{(k)}X)(v)\text{ with }\\
    (T_\bbW^{(k)}X)(v)=\int_0^1 \bbW(u,v)(T_\bbW^{(k-1)}X)(u)du
\end{align}
where $X,Y$ are graphon signals, $*\bbW$ defines the convolution operator, and $T^{(0)}=\bbI$ is the identity\cite{ruiz2020graphonFourierTransform}. Projecting the filter \eqref{eqn:graphon_filter} onto the eigenbasis $\{ \phi_i\}_{i\in\integers \smallsetminus 0}$, the graphon convolution admits a spectral representation given by,
\begin{align}
h(\lambda)= \sum_{k=0}^{K-1} h_k \lambda^k.
\end{align}
Like its graphon counterpart (cf. \eqref{eqn:graph_spectral}) the graphon convolution admits a spectral representation that only depends on the coefficients of the filter $\bbh$. 

Graphon Neural Networks (WNNs) are the extension of GNNs to graphon data. Each layer of a WNN is composed by a graphon convolution \eqref{eqn:graphon_filter}, and a non-linearity $\rho$. Denoting $F_l$ the features at layer $l$, we can group the parameters of the $F_{l-1}\times F_l$ convolutions into $K$ matrices $\{\bbH_{lk}\}\in\reals^{F_{l-1}\times F_l}$, to write the $f$
th feature of the $l$th layer as follows, 
\begin{align}\label{eqn:WNN_layer_wise}
	X_l^f=\rho \bigg ( \sum_{g=1}^{F_{l-1}} \sum_{k=1}^{K-1} (T_\bbW ^{(k)} X_{l-1}^g)[\bbH_{lk}]_{gf}  \bigg)
\end{align}
for $1\leq g\leq F_0 $. For an $L$ layered WNN, $X_0^g$ is given by the input data $X^g$, and \eqref{eqn:WNN_layer_wise} is repeated for $1\leq l \leq L$, and the output of the WNN is given by $Y=X_L$. Analogoues to GNNS, a more succint representation of the WNN  can be obtained be grouping all the coefficients, i.e. $\phi(X;\ccalH,\bbW)$, with $\ccalH=\{\bbH_{lk}\}_{lk}, 1\leq l \leq L$, and $0\leq k \leq K-1$.

\subsection{From Graphons to Graphs, and Back}
In this paper we focus on graphons, and graphon signals as generative models of graphs, and graph signals. Let $\{u_i \}_{i=1}^N$ be $n$ points sampled independently at uniformly from $[0,1]$, $u_i\sim\text{unif}(0,1)$. The $n$-node stochastic GSO of graph $\bbG_n$ is obtained from graphon $\bbW$ as follows, 
\begin{align}\label{eqn:sample_graph_bernoulli}
    [\bbS_n]_{ij}=[\bbS_n]_{ji}\sim\text{Bernoulli}(\bbW(u_i,u_j)))
\end{align}
A graph signal $\bbx$ can be obtained by evaluating the graphon signal, 
\begin{align}
	[\bbx]_i= X(u_i) \text{ for all }i\in[n]. \label{eqn:sample_graph_signal}
\end{align}
Therefore, we can obtain stochastic graphs $\bbS_n$, and graph signals $\bbx$  from graphon data $\bbW,X$. 

A graphon can be induced by a graph even if the node labels $\{u_i\}_{i=1}^N$ are unknown. Let $\bbS_n \in \reals^{n\times n}$ be the GSO of a graph, and define $u_i = (i-1)/n$ for $1\leq i\leq n$. We construct the intervals $I_i=[u_i, u_{i+1}]$ for $1\leq i \leq n$. Letting $\bbone$ be the indicator function, the induced graphon $\bbW_\bbS $, and graph signal $\bbx$ can be obtained as, 

\begin{align}\label{eqn:induced_graphon}
 \bbW_\bbS (u,v) &= \sum_{i=1}^{n}\sum_{j=1}^{n} [\bbS_n]_{ij} \bbone(u\in I_i)  \bbone(v\in I_j) ,\text{ and }\\
 X_n(u) &= \sum_{i=1}^n [\bbx_n]_i \bbone (u\in I_i).
\end{align}

\begin{algorithm}[t]
	\caption{Learning by transference}
	\label{alg:WNNL}
	\begin{algorithmic}[1]
		\State   Initialize $\ccalH_0,n_0$ and sample graph $\bbG_{n_0}$ from graphon $\bbW$
		\Repeat \ for epochs $0,1,\ldots$
		\For {$k$ =1,\dots, $|\ccalD|$}
		\State Obtain sample $(Y,X)\sim \ccalD$
		\State Construct graph signals $\bby_n,\bbx_n$ [cf. \eqref{eqn:sample_graph_signal}]
		\State Take learning step:\\ \quad\quad\quad\quad$\ccalH_{k+1}=\ccalH_{k}-\eta_k\nabla \ell(\bby_n,\bbPhi(\bbx_n;\ccalH_k,\bbS_n))$
		\EndFor
		\State {Increase number of nodes $n$}
		\State {Sample points uniformly $u_i\sim unif(0,1)$, $i=[1,\dots,n]$  }
		\State {Sample $\bbS_n$ Bernoulli from graphon $\bbW$ [cf. \eqref{eqn:sample_graph_bernoulli}]}
		\Until  convergence
	\end{algorithmic}
\end{algorithm}

\section{Learning By Transference}

%In this paper, 
We are interested in solving a statistical learning problem where the data is supported in very large graphs. In the limit, this corresponds to a graphon, so we consider the problem of learning a WNN.

Let $\ell:\reals\times\reals\to\reals$ be a non-negative loss function such that $\ell(x,y)=0$ if and only if $x=y$, and let $p(X,Y)$ be an unknown probability distribution over the space of the graphon signal. The graphon statistical learning problem is defined as
\begin{align}\label{eqn:graphon_srm}
 \min_{\ccalH} \mbE_{p(X,Y)} [\ell (Y,\phi(X;\ccalH,\bbW))].
\end{align}
Given that the joint probability distribution $p(X,Y)$ is unknown, a solution of \ref{eqn:graphon_srm} cannot be derived in close form but, under the learning paradigm, we assume that we have access to samples of the distribution $\ccalD = \{(X^j,Y^j)\sim p(X,Y),j=1,\dots,|\ccalD|\}$. Provided that the samples in $\ccalD$ are obtained independently, and that $|\ccalD|$ is sufficiently large, \ref{eqn:graphon_srm} can be approximated by its empirical version 
\begin{align}\label{eqn:graphon_erm}
	\min_{\ccalH} \frac{1}{|\ccalD|}\sum_{j=1}^{|\ccalD|} \ell (Y^j, \phi(X^j;\ccalH,\bbW)).
\end{align}
Like most other empirical risk minimization problems, the graphon empirical learning problem \eqref{eqn:graphon_erm} is solved in an iterative fashion by gradient descent. The $k$th gradient descent iteration is given by
\begin{align}\label{eqn:graphon_grad_descent}
	\ccalH_{k+1}=\ccalH_{k}-\eta_k\frac{1}{|\ccalD|}\sum_{j=1}^{|\ccalD|} \nabla_\ccalH \ell (Y^j,\phi(X^j;\ccalH_k,\bbW))
\end{align}
where $\eta_k\in(0,1)$ is the step size at iteration $k$.

In practice, however, the limit graphon $\bbW$ is either unknown or hard to measure, and step \eqref{eqn:graphon_grad_descent} cannot be computed. %However, given a number of nodes $n$, we can sample 
But this can be addressed by noting that this gradient can be \textit{approximated} by sampling graphs $\bbS_n$ (cf. \eqref{eqn:graphon_grad_descent}) and computing the gradient descent iteration on the graph. Explicitly, 
\begin{align}\label{eqn:graph_grad_descent}
	\ccalH_{k+1}=\ccalH_{k}-\eta_k\frac{1}{|\ccalD|}\sum_{j=1}^{|\ccalD|} \nabla_\ccalH \ell (\bby^j,\phi(\bbx^j;\ccalH_k,\bbS_n)).
\end{align}
To show that the graphon \eqref{eqn:graphon_grad_descent} and the graph gradient descent iterations \eqref{eqn:graph_grad_descent} are close, we need the following assumptions.
\begin{definition}[Lipschitz Functions] \label{def:norm_lips}
	A function $f$ is $A$-Lipschitz on the variables $u_1, \ldots, u_d$ if it satisfies $|f(v_1,v_2,\ldots,v_d)-f(u_1,u_2,\ldots,u_d)| \leq A\sum_{i=1}^d |v_i-u_i|$.
	If $A=1$, we say that this function is normalized Lipschitz.
\end{definition}
\begin{definition}[Node Stochasticity]\label{def:node_sto}
For a fixed probability $\ccalX_1\in[0,1]$, the node stochasticity constant on $n$ nodes, denoted $\alpha(\ccalX_1,n)$ is defined as $\alpha(\ccalX_1,n)=\log((n+1)^2/\log(1-\ccalX_1)^{-1})$.
 \end{definition}
\begin{definition}[Edge Stochasticity]\label{def:edge_sto}
	For a fixed probability $\ccalX_2\in[0,1]$, the edge stochasticity constant on $n$ nodes, denoted $\beta(\ccalX_2,n)$ is defined as $\beta(\ccalX_2,n)=\sqrt{n\log(2n/\ccalX_2)}$.
\end{definition}
\begin{assumption} \label{as1}
	The graphon $\bbW$ and graphon signals $X,Y$ are normalized Lipschitz.
\end{assumption} 
\begin{assumption} \label{as2}
	The convolutional filters $h$ are normalized Lipschitz and non-amplifying, i.e., $\|h(\lambda)\|<1$.
\end{assumption} 
\begin{assumption} \label{as3}
	The activation functions and their gradients are normalized Lipschitz, and $\rho(0)=0$.
\end{assumption} 
\begin{assumption} \label{as4}
	The loss function $\ell: \reals\times\reals\to\reals^{+}$ and its gradient are normalized Lipschitz, and $\ell(x,x)=0$.
\end{assumption}
\begin{assumption} \label{as5}
	For a fixed value of $\ccalX_3 \in (0,1)$, $n$ is such that $n-{\log (2n /\ccalX_3)}/{d_\bbW}> {2}/{d_\bbW}$ where $d_\bbW$ denotes the maximum degree of the graphon $\bbW$, i.e., $d_\bbW=\max_v \int^1_0 \bbW(u,v)du$.
\end{assumption} 
% The gradient approximation result is formally stated in Theorem \ref{thm:grad_Approximation}.
\begin{theorem}\label{thm:grad_Approximation}
	%(Graph to Graphon Learning Step Approximation).
	Consider the ERM problem in \eqref{eqn:graphon_erm} and let $\bbPhi(X;\ccalH,\bbW)$ be an $L$-layer WNN with $F_0=F_L=1$, and $F_{l}=F$ for $1\leq l \leq L-1$.  Let $c \in (0,1]$ and assume that the graphon convolutions in all layers of this WNN have $K$ filter taps [cf. \eqref{eqn:graphon_filter}]. Let $\bbPhi(\bbx_n;\ccalH,\bbS_n)$ be a GNN sampled from $\bbPhi(X;\ccalH,\bbW)$ as in \eqref{eqn:sample_graph_bernoulli}, and \eqref{eqn:sample_graph_signal}. Under assumptions AS\ref{as1}--AS\ref{as5}, it holds that
	\begin{align}\label{eq:thm_1}
	\mbE[\|\nabla_{\ccalH}&\ell(Y,\phi(X;\ccalH,\bbW))-\nabla_{\ccalH}\ell(Y_n,\phi(X_n;\ccalH,\bbW_n))\|]\nonumber\\
	&\leq \gamma c+\ccalO\bigg( \max(\alpha(1/\sqrt{n},n),\beta(1/\sqrt{n},n))\bigg)
	%3L^2F^{2L-2}\bigg( 1+\frac{\pi B^{c}_{\bbW \bbW_n}}{\delta_{\bbW \bbW_n}^c}\bigg)\sqrt{\frac{\log (2n^{3/2})}{n}}+12L^2F^{2L-2}c \label{eq:thm_1}
	\end{align}
	where $Y_n$ is the graphon signal induced by $[\bby_n]_i = Y(u_i)$ [cf. \eqref{eqn:induced_graphon}], and $\gamma=12\sqrt{K^5 F^{5L-5}}$. %The fixed constants $B^c_{\bbW}$ and $\delta^c_{\bbW \bbW_n}$ are the $c$-band cardinality and the $c$-eigenvalue margin of $\bbW$ and $\bbW_n$ respectively [cf. Definitions \ref{def:c_band_cardinality},\ref{def:c_eigenvalue_margin} in the supplementary material].
\end{theorem}

%For a given input-output pair $(X^j,Y^j)\in\ccalD$, Theorem \ref{thm:grad_Approximation} upper bounds the distance between the learning steps on the graph and on the graphon as a function of the number of nodes. 
As expected, the bound in \eqref{eq:thm_1} decreases with the number of nodes, but it also contains a constant term called the \textit{non-transferable} bound, which corresponds to high-frequency spectral components that converge more slowly with $n$ (see \cite[Sec. IV.C]{ruiz2021transferability}).

\begin{figure*}[ht]
     \centering
     \begin{subfigure}[b]{0.45\textwidth}
         \centering
         \includegraphics[width=\textwidth]{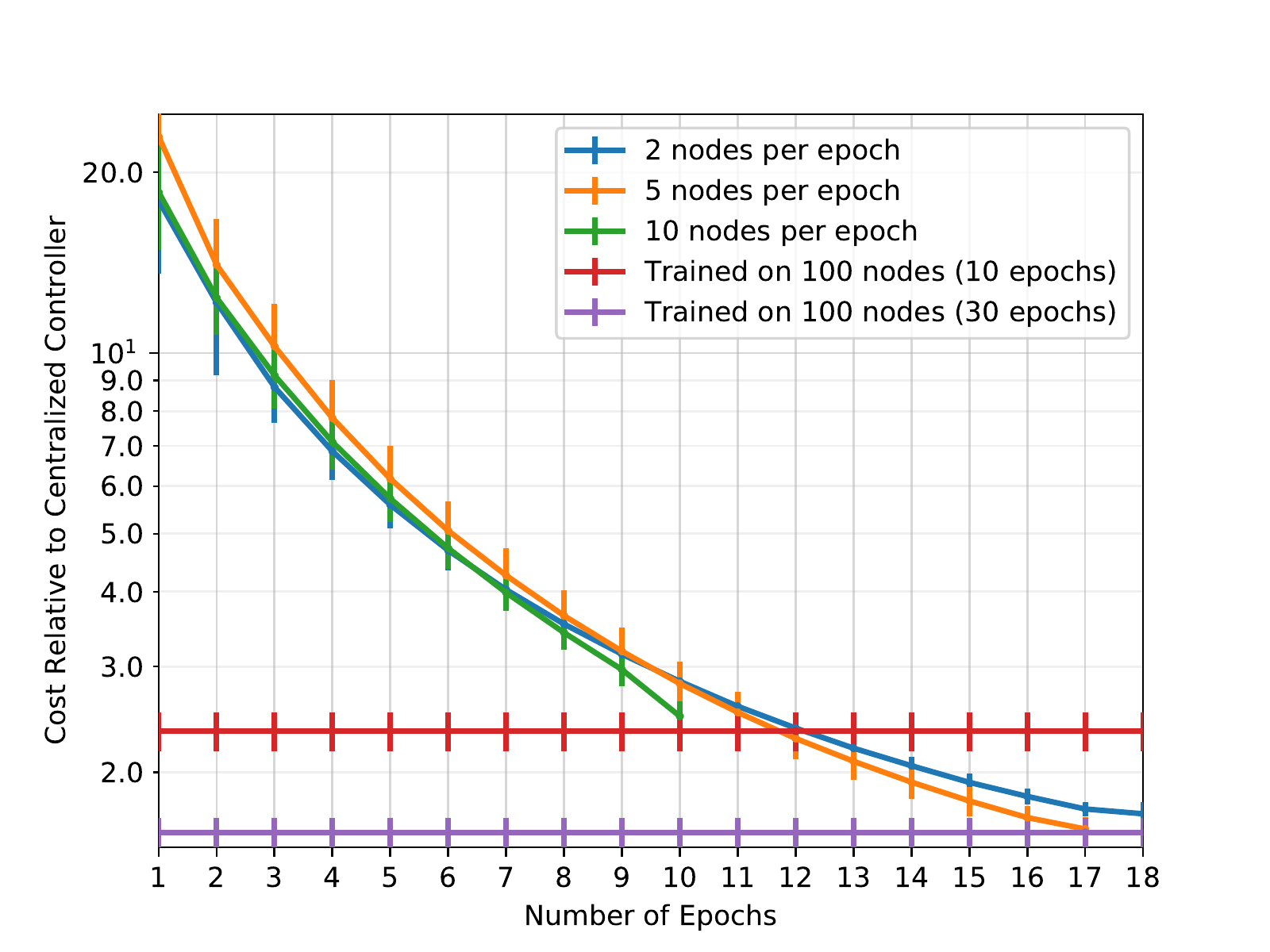}
         \caption{Starting with $10$ nodes}
     \end{subfigure}
     \hfill
     \begin{subfigure}[b]{0.45\textwidth}
         \centering
         \includegraphics[width=\textwidth]{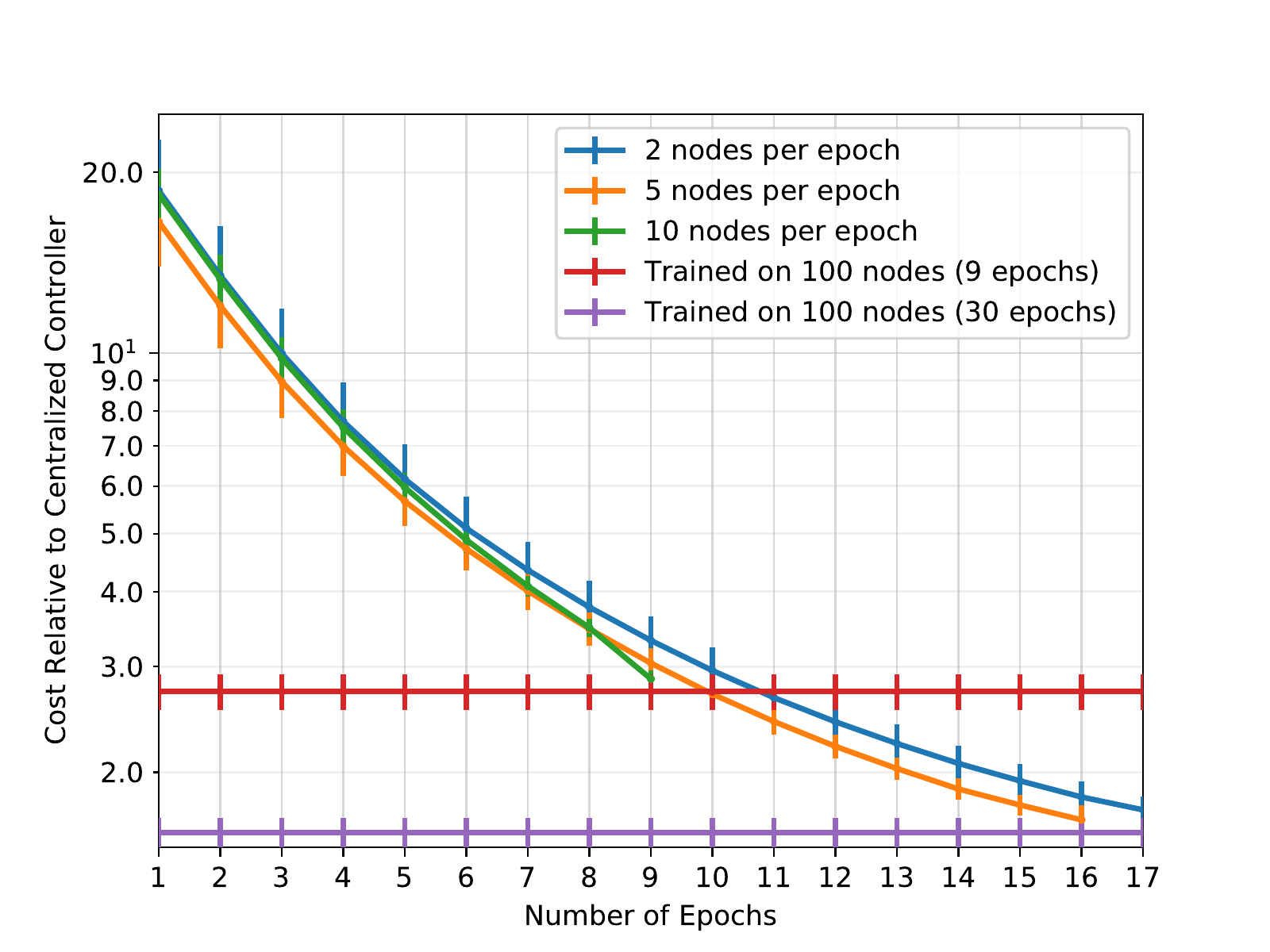}
         \caption{Starting with $20$ nodes}
     \end{subfigure}
	\caption{Velocity variation of the flocking problem for the whole trajectory in the testing set relative to the centralized controller.}
	\label{fig:flocking}
\end{figure*}

\subsection{Algorithm Construction}
Algorithm \ref{alg:WNNL} presents a simple strategy for training GNNs for large graphs: increasing the number of nodes and resampling the graph at regular intervals during the training process. The insight is that at the beginning of the learning process, the gradient on the graphon is large, given that the parameters $\ccalH_k$ are distant from the optimal values. As we train, the norm of the gradient on the graphon decreases, and we require a larger graph, i.e., a better approximation of the graphon, to follow the right learning direction on the graph. 

The advantage of Algorithm \ref{alg:WNNL} versus training on the large $N$-node graph directly is that given that the convolutions implement matrix-vector multiplications, this would require $\ccalO(N^2)$ computations. By implementing Algorithm \ref{alg:WNNL}, we are able to bring this down to $\ccalO(n^2)$ computations, $n<N$, without compromising optimality. What is more, Algorithm \ref{alg:WNNL} might not require evaluating gradients on the very large graph, as it could be the case that smaller graphs render the desired optimality conditions. 

% Algorithm \ref{thm:grad_Approximation} provides an intuitive result, given that the approximation bound depends on the number of nodes, it is possible to follow the learning direction on the graphon, if we keep the learning direction on the graph close. 

\subsection{Algorithm Convergence}
To show the convergence of Algorithm \ref{alg:WNNL}, we need additional Lipschitz assumptions.
\begin{assumption} \label{as7}
	The graphon neural network $ \bbPhi(X;\ccalH,\bbW)$ is $\AWNN$-Lipschitz, and its gradient $ \nabla_\ccalH\bbPhi(X;\ccalH,\bbW)$ is $\AgWNN$-Lipschitz, with respect to the parameters $\ccalH$ [cf. Definition \ref{def:norm_lips}].
\end{assumption}
\begin{theorem}\label{thm:WNN_learning}
	%Consider t$\sum \eta_k =\infty$,$\sum \eta_k^2 \leq\infty$. Under Assumptions \ref{as1}-\ref{as7}, if the loss $\ell(Y_n,\Phi(\bbW_n,X_n))$ converges to a finite value, and the.
	Consider the ERM problem in \eqref{eqn:graphon_erm} and let $\bbPhi(X;\ccalH,\bbW)$ be an $L$-layer WNN with $F_0=F_L=1$, and $F_{l}=F$ for $1\leq l \leq L-1$.  Let $c \in (0,1]$, $\epsilon>0$, step size $\eta<{\Anl}^{-1}$, with $\Anl=\AnWNN+\AWNN F^{2L} \sqrt{K}$ and assume that the graphon convolutions in all layers of this WNN have $K$ filter taps [cf. \eqref{eqn:graphon_filter}]. Let $\bbPhi(\bbx_n;\ccalH,\bbS_n)$ be a GNN sampled from $\bbPhi(X;\ccalH,\bbW)$ as in \eqref{eqn:induced_graphon}. Consider the iterates generated by equation \eqref{eqn:graph_grad_descent}, under Assumptions AS\ref{as1}-AS\ref{as7}, if at each step $k$ the number of nodes $n$ verifies 
	% 		\bal\label{eqn:thm2_condition}
	% 		C\|\nabla_{\ccalH}\ell(Y_n,\bbPhi(X_n;\ccalH_k,\bbW_n))  \|^{-1}\leq \sqrt{n}
	% 		\eal
	% 	\begin{align}\label{eqn:thm2_condition}
		%     &\gamma c +\ccalO\bigg( \sqrt{\frac{\log(n^{3/2})}{n}}\bigg)\nonumber \\
		%     &< \frac{1-\Anl\eta-\epsilon}{2} \|\nabla_{\ccalH} \ell(Y_n,\bbPhi(X_n;\ccalH_k,\bbWn)\|
		%     \end{align}
	%     then in finite time we will achieve an iterate $k^*$ such that the coefficients $\ccalH_{k^*}$ satisfy
	%     \begin{align}\label{eqn:them2_claim}
		%     \mbE_\ccalD[\|\nabla_{\ccalH}\ell(Y,\bbPhi(X;\ccalH_{k^*},\bbW))\|]\leq \gamma c+\epsilon, \quad 
		%     \end{align} 
	\begin{align}\label{eqn:thm2_condition}
		\mbE[\|\nabla_{\ccalH_{k}}&\ell(Y,\bbPhi(X;\ccalH_k,\bbW))-\nabla_{\ccalH}\ell(Y_n,\bbPhi(X_n;\ccalH,\bbW_n))\|]  \nonumber\\
		& +\epsilon < 	\|\nabla_{\ccalH} \ell(Y,\bbPhi(X;\ccalH_k,\bbW))\|%^2  /2 F^{2L} \sqrt{K}
	\end{align}
	then Algorithm \ref{alg:WNNL} converges to an $\epsilon$-neighborhood of the solution of the Graphon Learning problem \eqref{eqn:graphon_erm} in at most $k^*=\ccalO(1/\epsilon^2)$ iterations, with $\gamma=12\sqrt{K^5 F^{5L-5}}$].
\end{theorem}
Theorem \ref{thm:WNN_learning} provides the conditions under which Algorithm \ref{alg:WNNL} converges to a neighborhood of the first order stationary point of the empirical graphon learning problem \eqref{eqn:graphon_erm}. Other than the aforementioned smoothness assumptions, we only need to satisfy condition \eqref{eqn:thm2_condition} at every epoch, i.e., the norm of the gradient on the graphon has to be larger than the difference between the gradients on the graph and graphon. 
\section{Experiments}
\label{sec:Experiments}
We consider the problem of coordinating the velocity of a set of $n$ agents while avoiding collisions. At each time $t$ each agents knows its own position $r(t)_i\in\reals^2$, and speed $v(t)_i\in\reals^2$, and reciprocally exchanges it with its neighbors. Communication links $[\bbS]_{ij}$ exists if the distance between two agents $i,j$ is smaller that $2$ meters. At each time $t$ the controller sets an acceleration that remains constant over an interval of $T_s=20ms$. The system dynamics is governed by, 
\begin{align}
    r_i(t+1)&=u_i(t)T_s^2/2+v_i(t)T_s+r_i(t),\\
    v_i(t+1)&=u_i(t)T_s+v_i(t).
\end{align}
We define the velocity variation of the team as $\sigma_{\bbv(t)}=\sum_{i=1}^{n}\|v_i(t)-\bar \bbv(t)\|^2$, and the collision avoidance potential 
\begin{align*}
\begin{split}
CA_{ij}=\begin{cases}
	\frac{1}{\|r_i-r_j\|^2}-\log(\|r_i-r_j\|^2) & \text{if $\|r_i-r_j\| \leq R_{CA}$}\\
	\frac{1}{R_{CA}^2}-\log(R_{CA}^2) & \text{otherwise,}
\end{cases} 
\end{split}
\end{align*}
with $R_{CA}=1m$. We consider a centralized controller whose action is given by $u_i(t)^*=-n(v_i-\bar \bbv)+\sum_{j=1}^n\nabla_{r_i}CA(r_i,r_j)$ \cite{tanner2003stable}.

The learning setting is created by training a GNN that mimics the output of the centralized controller. To do so, we define the empirical risk minimization problem over the dataset $\ccalD=\{\bbu^*_{m},\bbx_m\}_m$ for $m\in[0,400]$,
\begin{align}
    \min_\ccalH \sum_{m=1}^{|\ccalD|}\|\bbu^*_{m}-\bbPhi(\bbx_m;\ccalH,\bbS)\|^2.
\end{align}
In Figure \ref{fig:flocking} we can see the velocity variation of the learned GNN measured on unseen data. Figure \ref{fig:flocking} validates the utility of the proposed method, as we are able to learn GNNs utilizing Algorithm \ref{alg:WNNL} that achieve a comparable performance with the GNN trained on all the nodes.
GNNs are that are trained with starting number of nodes $n_0=\{10,20\}$, and that add $10$ agents per epoch (green line) are able to achieve a similar performance when reaching $100$ agents than the one they would have achieved by training with $100$ agents the same number of epochs. This is the empirical manifestation of Theorem \ref{thm:WNN_learning}. Moreover, adding less agents per epoch (orange and blue lines) still achieve the same performance, but it takes more epochs to achieve. 
Overall, all the presented configurations are able to obtain a comparable performance that the one obtained with the full graph of $100$ agents while taking steps of graphs of growing sizes.

\section{Conclusion}
In this paper we presented a method for learning GNNs on very large graphs by growing the size of the graph as we train. Denoted learning by transference, we exploit the fact that the norm of the gradient on WNN decreases as it approaches a minima, and so we increase the precision at which we estimate it as epochs increase. We provide a proof of convergence of our algorithm, as well as numerical experiments on a multi-agent problem.

\newpage

\bibliographystyle{IEEEbib}
%\bibliography{strings,refs}
\bibliography{bib}
\newpage
\appendix
\section{Proof of Theorem 1}
Theorem \ref{thm:WNN_learning} follows directly from Theorem \cite[Theorem $1$]{cervino2021increase}, and  \cite[Lemma $3$]{cervino2021increase} with $\alpha$ as in the node stochasticity. It is also needs to be noted that $\ccalX_1$, $\ccalX_2$, and $\ccalX_3$ should be selected as $\sqrt{n}$. The maximum in the final bound comes from fact that the bound is governed by the maximum rate between $\beta$, and $\alpha$.

\section{Proof of Theorem 2}
See \cite[Theorem $2$]{cervino2021increase}.
% Denoting $P_\ccalC$ the primal problem \ref{eq:GraphRobustProblem} over the non-parametric class of functions $\ccalC$, by \cite[Proposition 2]{chamon2020probably}, we express know that, 
% \begin{align}
% D^* \leq P^*_\ccalC + (1+L\xi )\lambda^* \label{eqn:dualityGapLuiz}
% \end{align}
% Now noting that $\ccalP \subseteq \ccalC $, it implies that 
% \begin{align}
%  P^*_\ccalC \leq P^* \label{eqn:Primals}
% \end{align}
% Combining \eqref{eqn:dualityGapLuiz} and \eqref{eqn:GradientPrimal} attains the desired result. 
% \section{Proof of Theorem 2}
% See \cite[Theorem 3]{chamon2020probably}. 

\end{document}